\documentclass{article}

\usepackage{PRIMEarxiv}

\usepackage[utf8]{inputenc} % allow utf-8 input
\usepackage[T1]{fontenc}    % use 8-bit T1 fonts
\usepackage{hyperref}       % hyperlinks
\usepackage{url}            % simple URL typesetting
\usepackage{booktabs}       % professional-quality tables
\usepackage{amsfonts,amsmath}       % blackboard math symbols
\usepackage{nicefrac}       % compact symbols for 1/2, etc.
\usepackage{microtype}      % micro typography
\usepackage{lipsum}
\usepackage{fancyhdr}       % header
\usepackage{graphicx}       % graphics
\graphicspath{{media/}}     % organize your images and other figures under media/ folder

\usepackage{amsmath}
\usepackage{algorithm}
\usepackage{algpseudocode}
\title{Error-Centric PID Untrained Neural-Net (EC-PIDUNN) For Nonlinear Robotics Control
}

\author{
  Waleed Razzaq \\
  School of Intelligent Science $\&$ Engineering \\
  University of Science $\&$ Technology of China (USTC) \\
  Hefei, Anhui \\
  China\\
  \texttt{waleedrazzaq@mail.ustc.edu.cn} \\
}
\begin{document}
\maketitle

\begin{abstract}
Classical Proportional-Integral-Derivative (PID) control has been widely successful across various industrial systems such as chemical processes, robotics, and power systems. However, as these systems evolved, the increase in the nonlinear dynamics and the complexity of interconnected variables have posed challenges that classical PID cannot effectively handle, often leading to instability, overshooting, or prolonged settling times. Researchers have proposed PIDNN models that combine the function approximation capabilities of neural networks with PID control to tackle these nonlinear challenges. However, these models require extensive, highly refined training data and have significant computational costs, making them less favorable for real-world applications. In this paper, We propose a novel EC-PIDUNN architecture, which integrates an untrained neural network with an improved PID controller, incorporating a stabilizing factor (\(\tau\)) to generate the control signal. Like classical PID, our architecture uses the steady-state error \(e_t\) as input bypassing the need for explicit knowledge of the systems dynamics. By forming an input vector from \(e_t\) within the neural network, we increase the dimensionality of input allowing for richer data representation. Additionally, we introduce a vector of parameters \( \rho_t \) to shape the output trajectory and a \textit{dynamic compute} function to adjust the PID coefficients from predefined values. We validate the effectiveness of EC-PIDUNN on multiple nonlinear robotics applications: (1) nonlinear unmanned ground vehicle systems that represent the Ackermann steering mechanism and kinematics control, (2) Pan-Tilt movement system. In both tests, it outperforms classical PID in convergence and stability achieving a nearly critically damped response.

\end{abstract}

% keywords can be removed

\section{Introduction}
All real-world industrial systems exhibit significant nonlinear behavior as their complexity increase. In applications such as unmanned vehicles \cite{dagher2014design, morel2009applied, razzaq2023neural}, robotics \cite{sun2020reduced, yang2024deterministic}, and power systems \cite{shi2019perturbation, chen2024adaptive} the presence of multiple inter-connected control variables often lead to dynamic interactions that are difficult to control with traditional control strategies. Classical Proportional-integral-Derivative (PID) Controllers remain one of the most widely used control strategies largely because they are model-free and have been successful in linear or locally linearized tasks \cite{yang2024deterministic}. Some research and reviews suggest that 85 to 90 $\%$ of industrial processes still utilize PID control primarily due to its simplicity and robustness in applications where the system dynamics are well understood or linearizable. However, the performance of PID controllers diminishes significantly when applied to highly nonlinear systems, particularly in environments where system dynamics are unpredictable or contain multiple interconnected time-variant variables. Traditionally PID parameters are tuned empirically often relying on trial-and-error approaches or simplified models of the system. In nonlinear environments, PID can lead to significant control errors such as instability or oscillatory behavior. In response to these challenges, several advanced control strategies have been developed including adaptive control \cite{seto1994adaptive}, fuzzy logic control \cite{wang1996approach}, neural network-based controllers \cite{polycarpou1991identification}, and decoupling control \cite{kalayciouglu2018frf}. These approaches aim to bypass the need for precise system modeling by dynamically adjusting control parameters or incorporating intelligent systems to handle nonlinearities. The end-to-end \cite{saltzer1984end} neural network technique is particularly important because treats the system holistically by directly mapping inputs to control commands but the main concern with this is that it does not guarantee adherence to the required dynamics.

Several hybrid control strategies have also been developed to optimize the tuning of PID controllers each leveraging the strength of various techniques such as neural networks \cite{zeng2019adaptive}, fuzzy logic \cite{tang2001optimal}, and evolutionary computation \cite{iruthayarajan2009evolutionary}. Among these, PID Neural Networks (PIDNN) stand out for their ability to combine the function approximation capabilities of neural networks with the efficiency of classical PID control. Kang et.al \cite{kang2014adaptive} introduced a PIDNN framework where the Particle swarm Optimization algorithm is used to initialize the neural network weights. The PSO effectively optimizes weights but often suffers from slow convergence and high computational cost, particularly when applied to large-scale systems or real-time applications. Cong et al. \cite{cong2005novel} focused on improving learning rate and stability in gradient-based PIDNNs and analyzed how different learning rates affect the closed-loop stability of PIDNN controllers, finding that faster convergence often comes at the cost of reduced robustness under parameter uncertainty. It suggests that while gradient-based methods can accelerate learning, they may require sophisticated mechanisms to ensure system stability during training. Ho et al. \cite{ho2006optimizing} formulated an optimization problem that utilizes fuzzy neural networks to fine-tune PID parameters for more efficient damping control. While this method significantly improves response time, its practical implementation is hindered by the need to predetermine numerous system parameters, making it difficult to scale across different applications. Recurrent neural networks (RNNs) have also been employed to streamline the PID tuning process as demonstrated by Kumar et al. \cite{kumar2014ann} The recurrent structure allows the network to model temporal dependencies which simplifies identification and controller implementation. However, this approach still requires a substantial amount of prior system knowledge and parameter tuning before the training process can begin, limiting its flexibility in real-time applications. Jian et al. \cite{jiang2021improved} introduced an enhanced firefly algorithm that dynamically adjusts PIDNN weights, demonstrating strong performance in aircraft engine control systems. Similarly, Hasan et al. \cite{hasan2023optimizing} employed a fusion of population extremal optimization and genetic algorithms to refine PIDNN tuning further. Yang et al. \cite{yang2024structure} proposed a decoupling strategy for nonlinear multivariable systems, addressing the strong coupling that often arises in these environments. While promising this approach requires further validation in real-world scenarios to assess its effectiveness. PIDNN has been successfully applied in robotics fields, demonstrating its adaptability to different control tasks. Kumar et al. \cite{kumar2014ann} implemented an RNN-based adaptive PID controller to tackle the complexities of nonlinear robotics. By adjusting the PID gains in real-time based on the system’s dynamic feedback, the controller significantly improved the task accuracy in applications like robotics arm control where precision and responsiveness are critical. Zhang et al. \cite{zhang2021adaptive} extended this by proposing a double-layer backpropagation neural network for PID control in a legged robot. The first layer of the architecture learns the relations between control parameters and system performance, while the second dynamically adjusts these parameters in response to changing environmental conditions. The two-tier approach allows the controller to adapt to varying terrains improving the robot's stability and maneuverability. 

The quality of training data significantly influences the effectiveness of the current PIDNN techniques; any irregularities in this data can impair the controller’s performance \cite{jain2020overview}. Typically, the training data is tailored to a specific problem, making it task-specific and not easily transferable to other problems. The classical PID controller is popular due to its lack of data dependency and the straightforward empirical tuning of its parameters. Considering these factors, we propose an architecture called EC-PIDUNN where the PID gain coefficients are adjusted based on the $e_t$ making it error-centric. We also introduce a vector parameter ($\rho_t$) that can be shaped to control the output solution trajectory for gain calculation in the \textit{dynamic update} function. Furthermore, we aim to configure the network so that the neural network does not require training like classical PID. Although untrained (random) neural network weights can approximate functions \cite{rosenfeld2019intriguing}, leaving the weights at their initial random values can result in unstable responses. To mitigate this, we introduced an improved PID by incorporating a stabilizing factor ($\tau$) in the computation to stabilize the output. We applied the proposed technique to various nonlinear robotics applications, Ackermann unmanned ground vehicles, and humanoid pan-tilt mechanisms.

The main contributions of this paper are as follows:
\begin{enumerate}
    \item Proposes a novel EC-PIDUNN architecture that generates control output based on steady-state error ($e_t$) and eliminates the need for training weights.
    \item Introduces a parameter vector ($\rho_t$) that can be adjusted to tune the controller gains.
    \item Introduces a \textit{dynamic compute} function, implemented as a neural network output layer, to compute the control signal.
    \item Introduced an improved PID controller in \textit{dynamic update} with stabilizing factor \(\tau\) to stabilize the control signal.
    \item Demonstrates the effectiveness of the proposed architecture in complex interconnected variables nonlinear system of the unmanned ground vehicle problem.
\end{enumerate}
The remainder of the paper is organized as follows: Section 2 discusses the theoretical aspects of classical PID and PIDNN, and introduces the foundational components of our architecture. Section 3 details the proposed technique and its algorithm. Section 4 applies this technique to multiple robotics applications including Ackermann unmanned vehicle and Pan-Tilt eye movement mechanism. The final section concludes the paper, suggesting potential improvements and future work.

\section{Preliminaries}

In this section, we are going to overview the building blocks of our proposed architecture.

\subsection{PID Control}
\begin{figure}[t]
\centering
\includegraphics[width=0.8\textwidth]{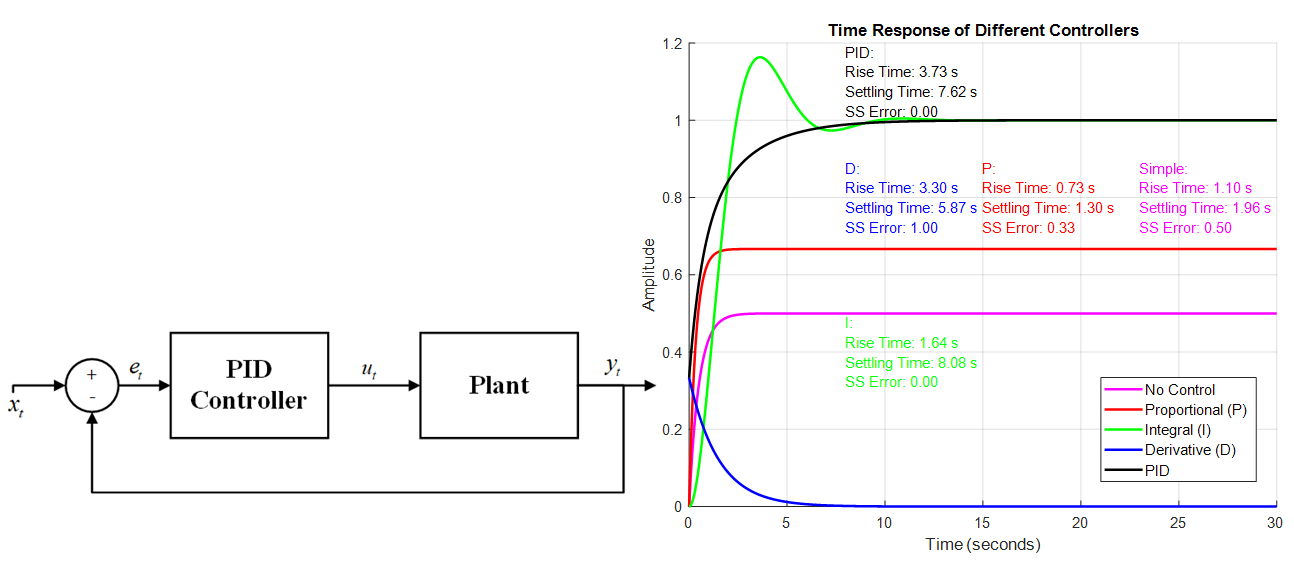}
\caption{\textbf{(a)} Classical-PID implementation on a plant process \textbf{(b)} Time-responses of controllers}
\label{fig1}
\end{figure}

A Proportional-Integral-Derivative (PID) controller is a feedback-based control loop mechanism \cite{van2004tuning} that is one of the most widely used methods for managing processes that require continuous control and automatic adjustment. The PID controller is highly popular due to its simplicity in both implementation and usage. Figure \ref{fig1}(a) shows a basic PID implementation for a plant process. Eqn.1 shows the mathematical computation of the PID control signal:
\begin{equation}
u_t = K_p e_t + K_i \int_0^T e_t \, dt + K_d \dot{e_t}
\end{equation}
where \(e_t\) is the continuous error signal at time \(t\), and \(K_p\), \(K_i\), and \(K_d\) represent the proportional, integral, and derivative gains respectively which determine the controller's response and \(u_t\) is the generated control signal. From a control theory perspective, the effects of each component of the controller on different aspects of the time response are illustrated in Figure \ref{fig1}(b) and summarized in Table \ref{tab1}.

%fig1

\begin{table}[ht]
    \centering
    \caption{Control System Response Characteristics}
    \begin{tabular}{@{}lcccc@{}} % Adjust column alignment here
        \toprule
        Controller & Rise Time (\(t_r\))    & Settling Time (\(t_s\))   & Overshoot (\(OS\))   & Steady-state Error (\(e_t\)) \\ 
        \midrule
        P (Proportional) & Decrease & Decrease   & Increases    & Decrease            \\
        I (Integral)     & Increase & Increase        & Decrease    & Eliminate          \\
        D (Derivative)   & Increase & Increase   & Reduce       & Increase          \\
        \bottomrule
    \end{tabular}
    \label{tab1}
\end{table}

\subsubsection{PID Neural-Net (PIDNN)}
\begin{figure}[h]
\centering
\includegraphics[width=0.6\textwidth]{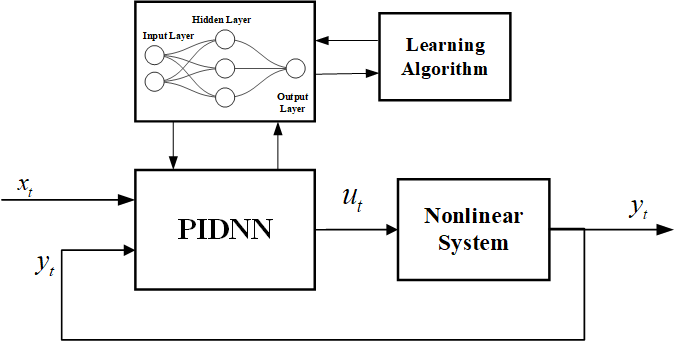}
\caption{PID Neural-Net architecture}
\label{fig2}
\end{figure}

A PIDNN architecture, in its simplest form, consists of a classical PID controller combined with a feedforward neural network. This architecture features a three-layer feedforward structure, which includes an input layer, a hidden layer to tune the gains, and an output layer that generates the control signal \cite{yongquan2003pid}. An architectural illustration can be seen in Figure \ref{fig2}. Here,  \(x_t\)\ is the system's input, \(y_t\) is feedback output and \(u_t\) is the control signal. Additionally, a learning algorithm, commonly backpropagation, is employed to train the neural network’s weights. The PIDNN operates by adjusting learning optimal PID parameters \(K_p, K_d, K_i\)\ as the system evolves or the neural network can also be trained directly to output \(u_t\) that mimics or enhances the traditional PID controller's behavior, making the system robust, adaptive and capable of handling complex dynamics.

\subsection{Multilayer Perceptrons}
A Multilayer Perceptron (MLP) is a type of artificial neural network commonly used for learning tasks such as classification, regression, and pattern recognition. It consists of a three-layer feedforward architecture containing an input layer, hidden layers, and an output layer. The input layer receives the input data, where each neuron in the input layer represents a feature or variable in the input data. This data is then passed to the hidden layers for processing before being forwarded to the output layer for decision-making. MLPs are widely known for their ability to approximate complex, nonlinear systems. They utilize activation functions such as \textit{ReLU}, \textit{Sigmoid}, or \textit{Tanh} in their hidden layers to introduce non-linearity, which allows them to capture complex patterns and approximate nonlinear functions. The single layer of an MLP system can be mathematically described as

\[
y = f\left(\sum_{i=1}^{n} w_i x_i + b\right)   
\]

where \( w_i \) is the weight of the neuron, \( x_i \) is the input, and \( b \) is the bias of the network.

%fig3
\begin{figure}[t]
\centering
\includegraphics[width=1.0\textwidth]{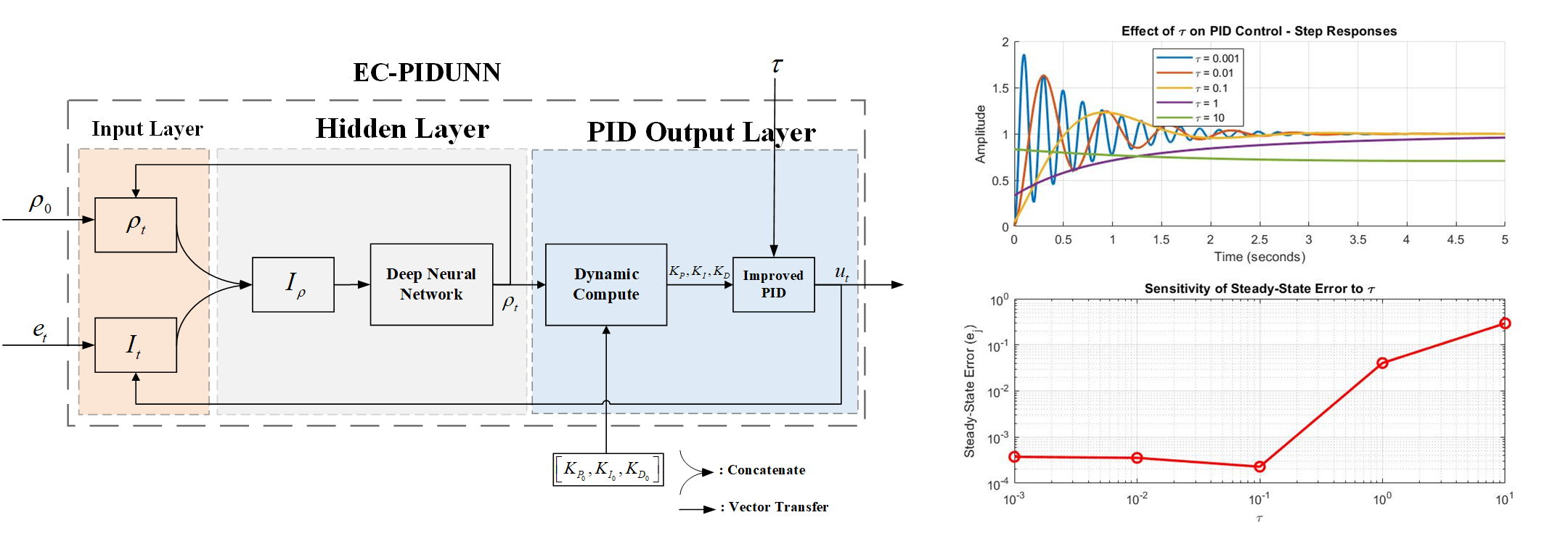}
\caption{\textbf{(a)} Internal architecture of EC-PIDUNN \textbf{(b) } Effect of \(\tau\) on \(e_t\)}
\label{fig3}
\end{figure}

\section{EC-PIDUNN}
The goal of this research is to modify and reconceptualize the basic architecture of PIDNN. Instead of using the system's object input \(x_t\) and feedback output \(y_t\), the input must be the steady-state error \(e_t\), essentially making the architecture error-centric. We introduced a vector of parameters \(\rho_t = [\rho_p, \rho_i, \rho_d] \in \mathbb{R}^3\) that can be shaped to tune the coefficients of the PID gains through an untrained (initial random weights) MLP-based neural network serving as function approximator. The input to the neural network is a vector \(I_{\rho} \in \mathbb{R}^6\), formed by combining \(I_t = [e_t, u_{t-1}, \Delta \epsilon_t] \in \mathbb{R}^3\), where \(e_t\), the previous control signal \(u_{t-1}\), the error difference \(\Delta \epsilon_t = e_t - u_{t-1} \), and \(\rho_t\), to increase dimensionality for rich data representation. Although an untrained neural network can generalize, its performance may degrade due to the randomness of the initial weights. To address this, we incorporate a feedback mechanism within the neural network, using \(u_t\) and \(\rho_t\) as inputs to constrain the network's hidden state. Additionally, we introduced an encapsulated \textit{dynamic compute} function in the output layer that dynamically calculates the PID coefficients based on \(\rho_t\), \(\Delta \epsilon_t\) and baseline gains \(K_{p0},K_{i0},K_{d0}\). Furthermore, an improved PID controller with stabilizing factor \(\tau\) is introduced in the output layer to calculate \(u_t\). The \(u_t\) along with the parameter vector \(\rho_t\), updates the \(I_t\) for the next iteration. This feedback mechanism ensures that the system converges to the correct solution and produces appropriate PID coefficients, regardless of the system's dynamics. The basic architecture of EC-PIDUNN is illustrated in Figure \ref{fig3}(a).

\subsection{Dynamic Compute}
The \textit{dynamic compute} function is introduced as the output layer of EC-PIDNN architecture to calculate and tune the PID coefficients based on a \(\Delta\epsilon_t\), \(p_t\), and base gains. Eqn 3 presents the mathematics to calculate PID gains. 
\begin{equation}
    K_i = K_{i0} + \rho_i \cdot \frac{\Delta \epsilon_t}{dt}
\end{equation}
\paragraph{Improved PID}In this configuration, depending on \(\rho_t\), the system may experience instabilities and randomness due to the untrained weights. To mitigate this effect, we introduce a stabilizing factor \(\tau\) to the integral and derivative controller. The equation for improved PID is presented. The effect of introducing \(\tau\) to the PID computation can be visualized in Figure \ref{fig3}. Algorithm 1 summarizes the EC-PIDUNN internal architecture.
\begin{equation}
        u(t) = K_p e_t + \frac{K_i}{\tau} \int_0^T e_t \, dt + \tau K_d \dot{e_t}
\end{equation}

%algorithm
\begin{algorithm}
\caption{EC-PIDUNN Algorithm}
\textbf{Require:} Steady-state error $e_t$, Neural Network $f(x, w) \to y$, Initial parameter vector $\rho_0$, Baseline gains $(K_{p0}, K_{i0}, K_{d0})$, damping factor $\tau$, Initial control signal $u_{t-1}$

\begin{algorithmic}[1]
\Procedure{EC-PIDUNN}{}
    \State Initialize vector $I_t = [e_t, u_{t-1}, \Delta \epsilon_t]$ and $\rho_t = [\rho_p, \rho_i, \rho_d]$
    \State \textbf{Input:} $e_t$, $u_{t-1}$, $\rho_0$
    \State Update $I_t$ and $\rho_t$
    \State Input vector: $I_\rho = \text{Concatenate}(I_t, \rho_t) \in \mathbb{R}^6$
    \State $\rho_t = \text{Neural Network}(I_\rho)$
    \State Update $\rho_t$
    
    \Procedure{Dynamic Compute}{$\rho_t$, $K_{p0}$, $K_{i0}$, $K_{d0}$}
        \State $K_p = K_{p0} + \rho_p \cdot \frac{\Delta \epsilon_t}{dt}$
        \State $K_i = K_{i0} + \rho_i \cdot \frac{\Delta \epsilon_t}{dt}$
        \State $K_d = K_{d0} + \rho_d \cdot \frac{\Delta \epsilon_t}{dt}$
        
        \Procedure{Improved PID}{$K_p$, $K_i$, $K_d$, $e_t$, $\tau$}
            \State $u_t = K_p e_t + \frac{K_i}{\tau} \int e_t \, dt + \tau \dot{e_t} K_d$
            \State \textbf{Output:} $u_t$
            \State $u_{t-1} = u_t$
        \EndProcedure
    \EndProcedure
\EndProcedure

\end{algorithmic}
\end{algorithm}

\section{Evaluation}
In this section, we will demonstrate the effectiveness of the proposed technique on multiple nonlinear robotic applications: (1) the Ackermann steering mechanism operating in a 2D plane under the influence of an applied force, and (2) a robotic pan-tilt mechanism.

%fig4
\begin{figure}[h]
\centering
\includegraphics[width=0.8\textwidth]{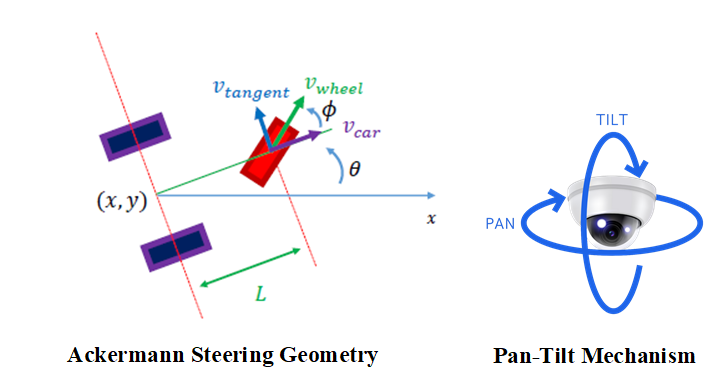}
\caption{\textbf{(a)} Ackermann Steering Geometry \textbf{(b)} Pan-Tilt mechanism }
\label{fig4}
\end{figure}

%fig5
\begin{figure}[t]
\centering
\includegraphics[width=0.5\textwidth]{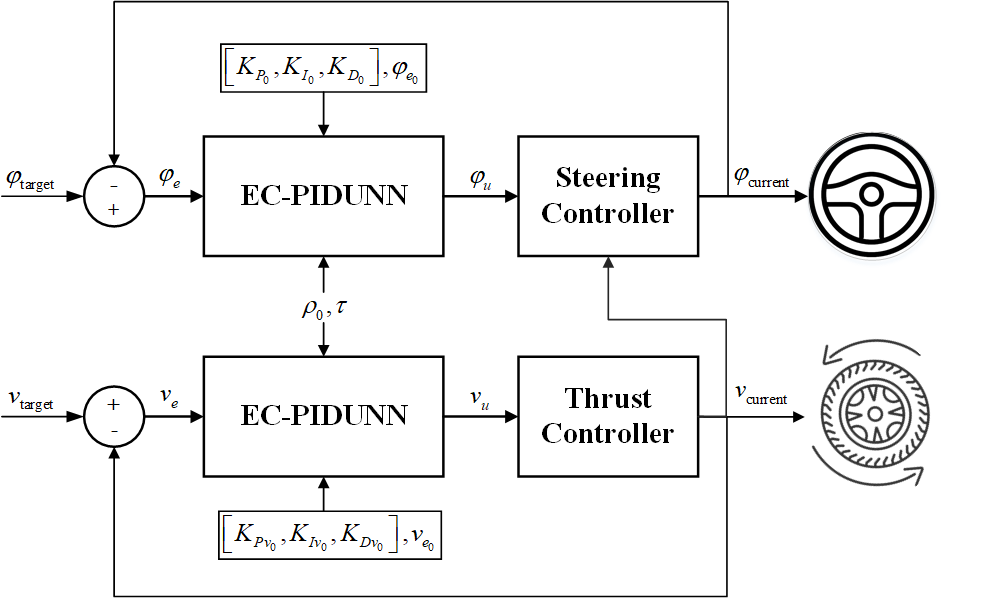}
\caption{Block Diagram of EC-PIDUNN implemented on vehicle system}
\label{fig5}
\end{figure}

\subsection{Ackermann Vehicle Dynamics}
\subsubsection{Ackermann Steering Geometry}
Ackermann's steering geometry is a fundamental concept in vehicle dynamics, particularly in the design of a car's steering mechanism. It ensures that all wheels follow circular paths when turning, reducing slip and improving handling. In a typical vehicle, the front wheels steer while the rear wheels provide forward motion. Figure \ref{fig4}(a) shows the demonstration of Ackermann geometry. A body-attached frame is used, with the x-axis pointing forward and the origin at the midpoint of the rear axle. When turning, the car's velocity aligns with the x-axis, and the speed is denoted as \(v_{\text{car}}\). Although the front wheels appear aligned, they trace different circular paths, with the outer wheels following a larger radius. To account for this, a virtual wheel is positioned midway between the front wheels, with its direction perpendicular to the line connecting the center of the turning circle. The angle \(\phi\) represents the steering angle between the car's x-axis and the direction of the virtual wheel. The car’s configuration is represented by \((x, y, \theta, \phi)\), where \(x, y\) are the car’s position, \(\theta\) is the orientation, and \(\phi\) is the steering angle. The system's two inputs are \(u_1\) and \(u_2\) \cite{mitchell2006analysis}.
\paragraph{Mathematical Modeling:}The differential kinematics for the car are specified by \(\dot{q} = f(q, u_1, u_2)\). The derivatives \(\dot{x}\) and \(\dot{y}\) can be found using the following equations:

\begin{equation}
\dot{x} = v_{\text{car}} \cos \theta 
\dot{y} = v_{\text{car}} \sin \theta 
\end{equation}

Additionally, the rate of change of the steering angle is given by:
\begin{equation}
\dot{\phi} = u_2 
\end{equation}

To calculate \(\dot{\theta}\), we refer to Figure 2 (right), where we find the following relationships for the car's turning:
\begin{equation}
\cos \phi = \frac{v_{\text{car}}}{v_{\text{wheel}}}
\end{equation}
\begin{equation}
\sin \phi = \frac{v_{\text{tangent}}}{v_{\text{wheel}}}
\end{equation}

Substituting the value of \(v_{\text{wheel}}\) from equation (6) into equation (7) yields:

\begin{equation}
 v_{\text{car}} \sin \phi = \frac{v_{\text{car}} \tan \phi}{\cos \phi} = v_{\text{tangent}}   
\end{equation}

We also observe that \(L \dot{\theta} = v_{\text{tangent}}\), which gives us:

\begin{equation}
\dot{\theta} = \frac{v_{\text{car}} \tan \phi}{L} = \frac{u_1 \tan \phi}{L}
\end{equation}

The state-space model of combined Ackermann steering control becomes
\begin{equation}
\begin{bmatrix}
\dot{x} \\
\dot{y} \\
\dot{\theta} \\
\dot{\phi}
\end{bmatrix}
=
\begin{bmatrix}
v_{\text{car}} \cos \theta \\
v_{\text{car}} \sin \theta \\
\frac{v_{\text{car}} \tan \phi}{L} \\
0
\end{bmatrix}
u_1
+
\begin{bmatrix}
0 \\
0\\
0\\
1
\end{bmatrix}
u_2
\end{equation}

\subsubsection{Kinematics of Vehicle}
In automotive systems, vehicle dynamics are influenced by several factors, including applied throttle force, vehicle mass, aerodynamic drag, and the capabilities of the powertrain. Understanding and controlling how these factors affect a car's motion is crucial for optimizing performance, safety, and energy efficiency. Let's consider a two-dimensional model of a car's motion. This model can be useful for understanding how forces influence the car's speed over time, while also accounting for real-world constraints, such as the maximum engine force and drag effects.
\paragraph{Mathematical Modeling:} Let’s assume a car with mass \( m \) is being pushed by a force \( F \) in a 2-D space and is subjected to aerodynamic drag, which is represented as:
\begin{equation}
F_{drag} = \frac{1}{2} C_d A \rho_{air} v_{car}^2
\end{equation}

where \( C_d \) is the drag coefficient, \( A \) is the reference area, \( \rho_{air} \) is the air density, and \( v_{car} \) is the velocity of the car.

The dynamic equation can be derived using Newton's second law:
\begin{equation}
\dot{v_{car}} = \frac{1}{m}(F - \frac{1}{2} C_d A \rho_{air} v_{car}^2)
\end{equation}

\subsection{Pan-Tilt Mechanism}
The pan-tilt mechanism is often employed in robotics systems for object tracking, surveillance, and visual servoing applications. A pan-tilt system provides two degrees of freedom, enabling precise control over the orientation of cameras or sensors. The mechanism can be modeled as two rotational joints with independent dynamics for pan (horizontal rotation about the vertical axis) and tilt (vertical rotation about the horizontal axis). Each axis of rotation is subject to its torque, inertia, and damping forces. Figure \ref{fig4}(b) shows the pan-tilt mechanism. Consider a pan-tilt system where $\theta_t$ is the pan-angle and $\phi_t$ is the tilt angle at time $t$. The dynamics of the system can be described by the second-order differential equations, which represent the angular motion.
\begin{equation}
I_{\theta} \ddot{\theta}(t) + b_{\theta} \dot{\theta}(t) = \tau_{\theta}(t)
\end{equation}

\begin{equation} 
I_{\phi} \ddot{\phi}(t) + b_{\phi} \dot{\phi}(t) = \tau_{\phi}(t)
\end{equation}

where \( I_{\theta} \) and \( I_{\phi} \) represent the moment of inertia of the pan and tilt joints, respectively. \( b_{\theta} \) and \( b_{\phi} \) are the damping coefficients for the pan-tilt motions, and \( \tau_{\theta}(t) \) and \( \tau_{\phi}(t) \) are the torques applied to the pan and tilt joints.

%fig6
\begin{figure}[t]
\centering
\includegraphics[width=0.9\textwidth]{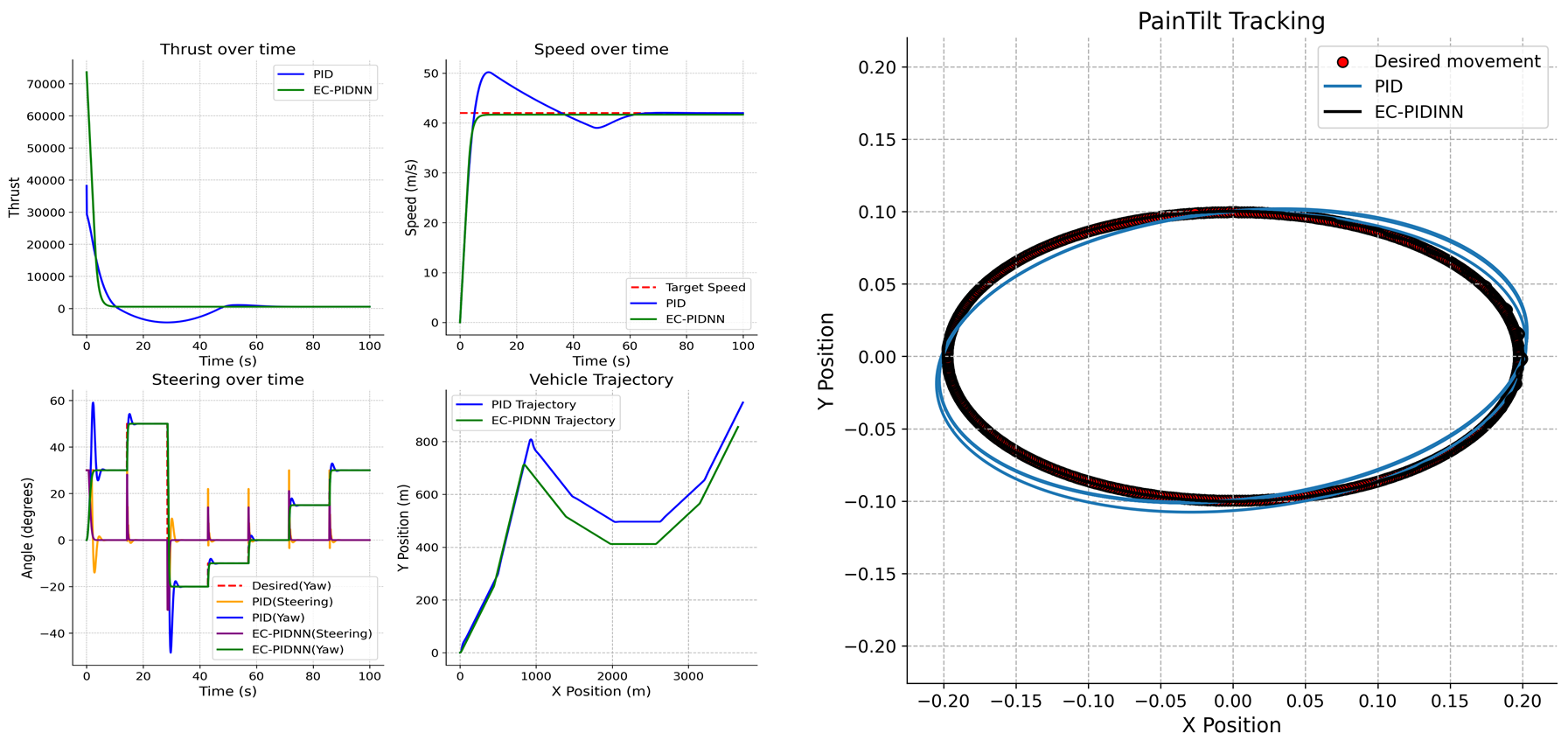}
\caption{\textbf{(a)} Vehicle system response \textbf{(b)} Pan-Tilt system response }
\label{fig6}
\end{figure}

\subsection{Simulation Results}
\subsubsection{Ackermann Vehicle Trajectory} 
The block diagram of the proposed technique is shown in Figure \ref{fig5}. In the upper feedback system, the Ackermann steering mechanism is controlled. The target yaw angle (\(\phi_{\text{target}}\)) is input into the adder/subtractor to calculate the error (\(\phi_e\)), which is then forwarded to the EC-PIDNN architecture, along with other parameters, to generate the steering signal (\(\phi_u\)) for the steering controller. This process continues until the desired target is achieved. The lower block follows a similar strategy to control the car's speed (\(v_{\text{car}}\)) to reach the desired speed (\(v_{\text{target}}\)). It is important to note that the initial vector of parameters (\(\rho_0\)) and the stabilizing factor (\(\tau\)) are shared between both implementations. Additionally, the current velocity (\(v_{\text{current}}\)) is also input into the steering controller for Ackermann control. Figure \ref{fig6}(a) presents the results of this implementation, comparing it with the traditional PID controller demonstrating that our proposed technique successfully achieved a nearly critically damped response with 0$\%$ overshoot. The time-response comparison is summarized in Table \ref{tab2}.

\begin{table}[h]
\centering
\caption{Comparison of controller performance metrics}
\begin{tabular}{lcccccc}
\toprule
\textbf{Controller} & \multicolumn{2}{c}{\textbf{Rise Time}} & \multicolumn{2}{c}{\textbf{Settling Time}} & \multicolumn{2}{c}{\textbf{Overshoot}} \\
\cmidrule(lr){2-3} \cmidrule(lr){4-5} \cmidrule(lr){6-7}
 & \textbf{Steering ($\phi$)} & \textbf{Speed ($v_{car}$)} & \textbf{Steering ($\phi$)} & \textbf{Speed ($v_{car}$)} & \textbf{Steering ($\phi$)} & \textbf{Speed ($v_{car}$)} \\
\midrule
Classical-PID & 9.8  & 6.71 & 14.5 & 5.6  & 66\%  & 19.4\% \\
EC-PIDUNN    & 9.65 & 6.65 & 1.9  & 3.7  & 0\%   & 0\% \\
\bottomrule
\end{tabular}
\label{tab2}
\end{table}

\subsubsection{Pan-Tilt Tracking} 
The objective of EC-PIDUNN is to make the pan-tilt mechanism track a moving target that follows a predefined trajectory in 2D space. Let the target’s position at time \( t \) be defined by the coordinates \( (x_t, y_t) \), where \( x_t \) corresponds to the horizontal axis and \( y_t \) corresponds to the vertical axis.

\begin{equation}
\theta_{\text{desired}}(t) = \arctan\left(\frac{x_t}{z}\right)
\end{equation}

\begin{equation}
\phi_{\text{desired}}(t) = \arctan\left(\frac{y_t}{z}\right)    
\end{equation}

Where \( z \) is the fixed distance from the camera to the target in the depth direction. This allows the pan-tilt system to compute the necessary angular rotations to keep the target centered in the field of view. The results of this test have been illustrated in Figure \ref{fig6}(b). The results suggest that EC-PIDUNN has successfully tracked the object with nearly critical response and near to zero overshoot while classical PID was unable to maintain its performance.

\section{Conclusion}
In this paper, we propose a novel error-centric PID untrained neural network (EC-PIDUNN) architecture that redefines the conventional PIDNN by incorporating an untrained (random-weighted) neural network as a function approximator alongside an improved PID controller with a stabilizing factor $\tau$. Rather than using the system’s desired input $x_t$ and feedback output $y_t$, our approach uses the steady-state-error $e_t$ as the system input, forming an input vector $I_{\rho}$. Additionally, we introduce a parameter vector $\rho_t$ that can be adjusted to shape the output trajectory in the hidden layer. In the output layer, we implement a \textit{dynamic update} function that tunes the PID coefficients based on $\rho_t$, $\Delta \epsilon_t$, and baseline gains, generating the control signal $u_t$ through the improved PID which is then fed back into the input layer for the next iteration thereby creating an internal feedback loop within the network. We applied this method to two well-known nonlinear robotics applications: (1) the vehicle control problem involving Ackermann steering and vehicle speed control under real-world aerodynamic disturbances and (2) the pan-tilt movement mechanism. Our results show that EC-PIDUNN consistently produces a near-critically damped response with near to 0\% overshoot and a fast settling time compared to traditional PID controllers. The rise time of EC-PIDUNN was occasionally equal to or but sometimes slower than that of classical PID controllers.

\paragraph{Future Work:}While EC-PIDUNN achieves superior performance compared to traditional PID, it comes with a computational cost and sensitivity to the neural network hyperparameters and stabilizing factor $\tau$. Additionally, the current version lacks noise and disturbance mitigation mechanisms. Developing a noise cancellation or filtering system could be a valuable direction for future research.

% \section*{Conflict of Interest}
% The authors declare that they have no conflicts of interest to disclose.
% \section*{Ethics Approval}
% This study was conducted in accordance with ethical standards.
% \section*{Funding}
% The research did not receive any funding from any organization.
% \section*{Data Availability}
% The data and code will be made public upon acceptance of the paper.
% \section*{Authors' Contribution}
% The corresponding author conducted all research and conceptualization.
% \section*{Acknowledgment}
% The authors wish to thank Prof. YunBo Zhao from the University of Science and Technology (USTC) for his support and valuable insights during this research.
% \section*{Human/Animal Participation}
% No human or animal participation involved in this research.

%Bibliography
\bibliographystyle{unsrt}  
\bibliography{references}

\end{document}